\documentclass[letterpaper]{article} 
\usepackage{aaai2026}  
\usepackage{times}  
\usepackage{helvet}  
\usepackage{courier}  
\usepackage[hyphens]{url}  
\usepackage{graphicx} 
\usepackage{natbib}  
\usepackage{caption} 
\usepackage{amsmath}
\usepackage{times}  
\usepackage{helvet}  
\usepackage{courier}  
\usepackage[hyphens]{url}  
\usepackage{graphicx} 
\usepackage{algorithm}
\usepackage{algorithmic}
\usepackage{bm}
\usepackage{amsbsy}
\usepackage{amssymb}
\usepackage{amsfonts}
\usepackage{amsmath}
\usepackage{newfloat}
\usepackage{listings}
\DeclareCaptionStyle{ruled}{labelfont=normalfont,labelsep=colon,strut=off} 
\floatstyle{ruled}
\newfloat{listing}{tb}{lst}{}
\floatname{listing}{Listing}
\title{Meta-Black-Box Optimization with Bi-Space Landscape Analysis \\and Dual-Control Mechanism for SAEA}
\author{
    Yukun Du\textsuperscript{\rm 1},
    Haiyue Yu\textsuperscript{\rm 1}\thanks{Corresponding author.},
    Xiaotong Xie\textsuperscript{\rm 1},
    Yan Zheng\textsuperscript{\rm 1},\\
    Lixin Zhan\textsuperscript{\rm 1},
    Yudong Du\textsuperscript{\rm 2},
    Chongshuang Hu\textsuperscript{\rm 1},
    Boxuan Wang\textsuperscript{\rm 1},
    Jiang Jiang\textsuperscript{\rm 1}
}
\affiliations{
    \textsuperscript{\rm 1}National University of Defense Technology,\\
    \textsuperscript{\rm 2}Xinxiang University,\\
    duyukun-nudt@outlook.com, yuhaiyue09@nudt.edu.cn, xiexiaotong20@nudt.edu.cn,\\ zhengyan24@nudt.edu.cn, zhanlixin98@outlook.com, duyudong668@outlook.com, \\huchongshuang@foxmail.com, boxuanwang24@nudt.edu.cn, jiangjiangnudt@nudt.edu.cn}

\usepackage{bibentry}

\begin{document}

\maketitle

\begin{abstract}
Surrogate-Assisted Evolutionary Algorithms (SAEAs) are widely used for expensive Black-Box Optimization. However, their reliance on rigid, manually designed components such as infill criteria and evolutionary strategies during the search process limits their flexibility across tasks. To address these limitations, we propose Dual-Control Bi-Space Surrogate-Assisted Evolutionary Algorithm (DB-SAEA), a Meta-Black-Box Optimization (MetaBBO) framework tailored for multi-objective problems. DB-SAEA learns a meta-policy that jointly regulates candidate generation and infill criterion selection, enabling dual control. The bi-space Exploratory Landscape Analysis (ELA) module in DB-SAEA adopts an attention-based architecture to capture optimization states from both true and surrogate evaluation spaces, while ensuring scalability across problem dimensions, population sizes, and objectives. Additionally, we integrate TabPFN as the surrogate model for accurate and efficient prediction with uncertainty estimation. The framework is trained via reinforcement learning, leveraging parallel sampling and centralized training to enhance efficiency and transferability across tasks. Experimental results demonstrate that DB-SAEA not only outperforms state-of-the-art baselines across diverse benchmarks, but also exhibits strong zero-shot transfer to unseen tasks with higher-dimensional settings. This work introduces the first MetaBBO framework with dual-level control over SAEAs and a bi-space ELA that captures surrogate model information.
\end{abstract}


\section{Introduction}

Surrogate-Assisted Evolutionary Algorithms (SAEAs) are effective approaches for solving expensive Black-Box Optimization (BBO) problems \cite{a3,a4}. By introducing surrogate models, such as Gaussian Processes (GPs), to approximate the true objective function and combining them with evolutionary algorithms for global search \cite{a23}, SAEAs can deliver high-quality solutions while reducing the number of expensive evaluations \cite{a2,a22}. However, existing SAEAs still rely on rigid, manually crafted components, such as infill criterion design and evolutionary strategy configuration, which limits their flexibility \cite{a6,a7,a21}. As a result, SAEAs struggle to dynamically adjust their components based on task characteristics and the evolving requirements of different search phases, thereby limiting their transferability and adaptability.

Beyond the fixed design of conventional SAEAs, Meta-Black-Box Optimization (MetaBBO) provides a more general, learning-based framework for adaptive optimization across diverse tasks and dynamic search conditions. By formulating algorithm configuration as a data-driven control process \cite{a8,a9}, MetaBBO reduces manual intervention and improves transferability and adaptability across tasks and search phases. As shown in Figure 1, it follows a bi-level structure \cite{a1}, where the meta-level gathers feedback from the low-level optimizer (i.e., BBO Method in Figure 1) via Exploratory Landscape Analysis (ELA) \cite{a10,a17,a18}. This feedback is then utilized by a meta-policy to dynamically adjust key components, such as optimizer selection and search strategies, thereby forming a closed-loop adaptation mechanism \cite{a11,a12,a13,a16}.

\begin{figure}[t]
\centering
\includegraphics[width=1.0\columnwidth]{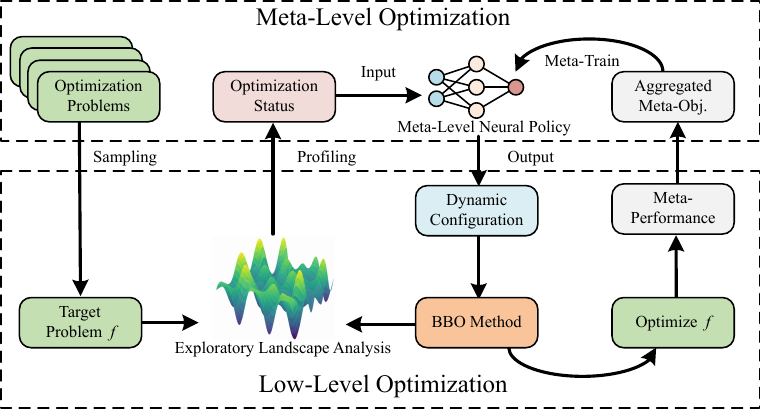}
\caption{The general workflow of MetaBBO}
\label{fig1}
\end{figure}

Despite the increasing attention to MetaBBO, most existing approaches still rely on traditional evolutionary algorithms as low-level optimizers \cite{a24,a25}. For instance, R2-RLMOEA \cite{a15} employs a reinforcement learning agent to switch among several evolutionary algorithms, while LDE \cite{a14} leverages an LSTM network to adjust key parameters—such as the mutation factor and crossover rate—under varying search conditions. Although these works highlight the potential of MetaBBO for adaptive control, the integration of SAEAs into the MetaBBO paradigm remains largely unexplored. This gap is primarily due to several fundamental challenges:

1) \textbf{Difficulty in landscape-aware representation}. In MetaBBO, ELA characterizes current search behavior via landscape-aware state representations that guide meta-policy decisions. However, constructing informative state representations remains difficult, especially under surrogate-assisted settings. The state should reflect both the population evaluated by the true objective and surrogate-derived signals such as predictive uncertainty, which are often high-dimensional, correlated, and hard to encode for effective learning and control.

2) \textbf{Lack of unified meta-level control over key components of SAEA}. Existing MetaBBO methods rarely investigate how to jointly control the key components of SAEA, including the surrogate model, evolutionary strategy, and infill criterion. For example, DRL-SAEA \cite{a20}, integrates SAEA as the low-level optimizer but restrict control to surrogate model management, without extending control to other essential modules. Designing a unified meta-policy that manages multiple tightly coupled modules and adapts to changing search conditions remains a challenging problem.

3) \textbf{Limitations of surrogate models}. Traditional surrogate models such as Gaussian Processes (GPs) suffer from high computational cost and poor scalability with dimensionality. In high-dimensional problems, they often produce unreliable predictions, limiting their ability to support meta-policy learning \cite{a19,a45}. Moreover, the high training cost of GPs makes them impractical for MetaBBO frameworks, which typically require frequent model updates across many tasks.

To promote the integration of SAEAs into the MetaBBO paradigm, we propose DB-SAEA, a novel MetaBBO framework for expensive multi-objective problems, which employs a SAEA as the low-level optimizer. DB-SAEA is designed to address the challenges by integrating the following techniques:

1) \textbf{Bi-space neural network-based ELA}. We design a bi-space ELA module using multiple Two-Stage Attention (Ts-Attn) blocks to extract landscape-aware features from both true and surrogate evaluation spaces. This design integrates surrogate model signals to construct optimization states. The resulting representations enable scalable landscape modeling across varying problem dimensions, population sizes, and numbers of objectives, and are jointly trained with the meta-policy to couple feature extraction with decision control.

2) \textbf{Dual-control mechanism for SAEA}. We propose a dual-control mechanism where the meta-policy jointly governs infill criterion selection and controls the evolutionary algorithm by deciding whether candidate solutions should undergo true evaluation. In each iteration, the evolutionary algorithm generates candidate solutions, and the meta-policy decides whether to proceed with true evaluation or continue surrogate-assisted search. If true evaluation is chosen, the meta-policy further selects the most suitable infill criterion based on the current optimization state.

3) \textbf{Efficient surrogate modeling with TabPFN}. To address the inefficiency and poor scalability with dimensionality of GPs, we adopt TabPFN \cite{a28}, a transformer-based probabilistic estimator tailored for tabular data. TabPFN supports fast inference without task-specific training, while providing accurate objective prediction and uncertainty estimation.

The main contributions of this work are as follows. 1) we are first to propose a dynamic, multi-objective landscape-aware representation method that is capable of capturing surrogate model information. 2) We overcome the limitations of single-component control by developing a dual-control mechanism, establishing a novel control paradigm for MetaBBO. 3) We integrate TabPFN into the SAEA framework as an efficient surrogate model for fast and accurate prediction with uncertainty estimation.

\begin{figure*}[t]
\centering
\includegraphics[width=1\textwidth]{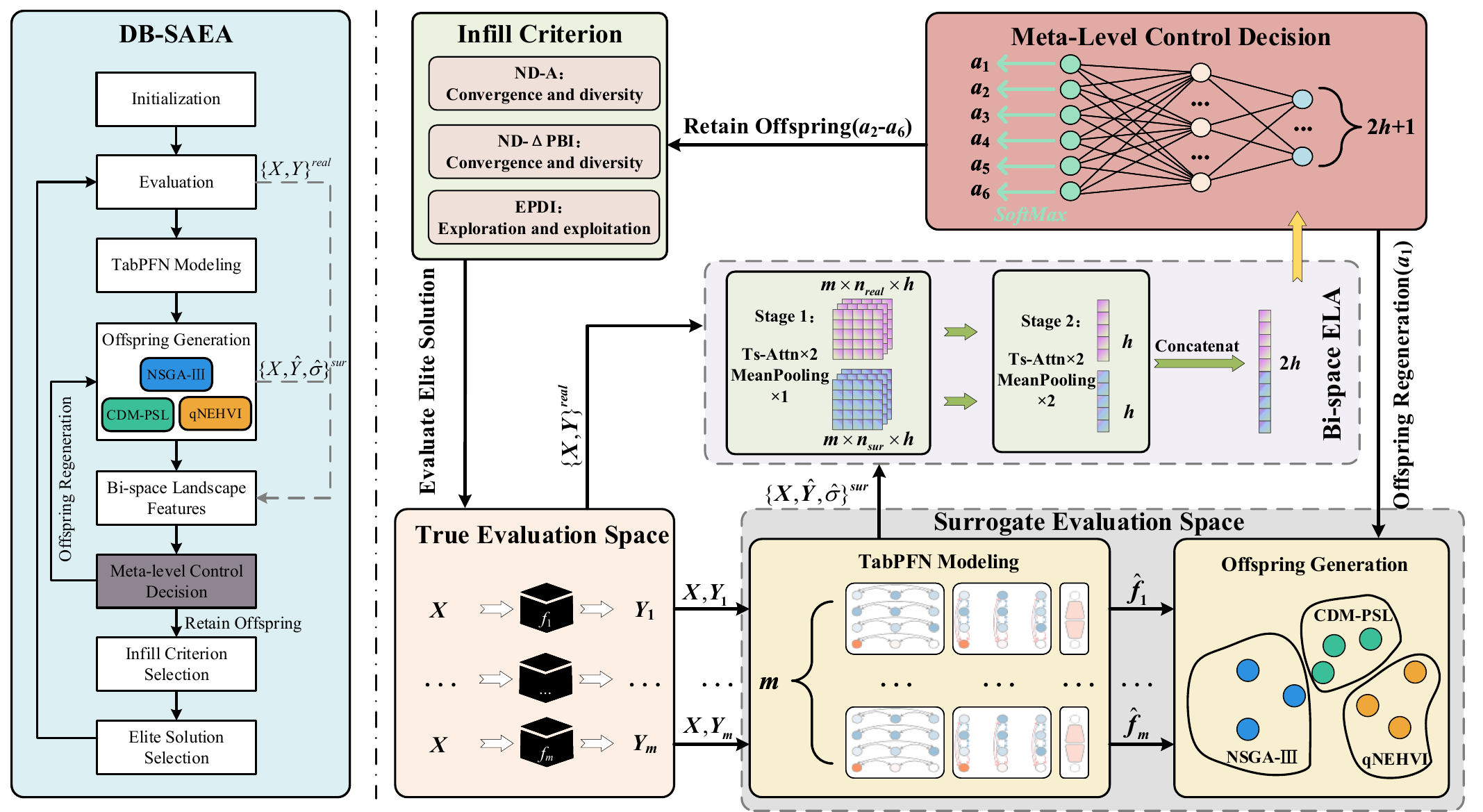} 
\caption{The framework of DB-SAEA. (right) DB-SAEA generates candidate solutions using hybrid sampling (NSGA-III, CDM-PSL, qNEHVI), evaluates them with TabPFNs, and extracts landscape-aware features via bi-space ELA. The meta-policy determines whether to perform true evaluation or continue surrogate-assisted search. Upon true evaluation, elite solutions are selected and $\mathcal{P}_\text{true}$ is updated.}
\label{fig2}
\end{figure*}

\section{Backgrounds}

\subsection{Neural Network-based ELA}

Recent advances in neural network-based ELA have introduced methods such as Deep-ELA \cite{a27} and NeurELA \cite{a26}, both leveraging attention mechanisms to model optimization landscapes. Deep-ELA, pre-trained on large-scale static problems, generates one-shot global representations but lacks adaptability to evolving optimization states and models only interactions across sample points. NeurELA mitigates these limitations by introducing a two-stage attention mechanism that captures both inter-solution and inter-dimension interactions, enhancing its representational capacity in dynamic MetaBBO tasks. However, it remains restricted to single-objective settings and, like Deep-ELA, does not incorporate feedback from surrogate models, limiting its ability to characterize the surrogate evaluation space.

To address the aforementioned limitations, we propose a bi-space neural network-based ELA tailored for multi-objective optimization, serving as a core component of the DB-SAEA framework. Building on existing neural ELA methods, our approach extends them by incorporating both true and surrogate evaluation spaces, enabling dynamic, landscape-aware representations across multiple objectives. Here, we construct the landscape representations of both the true and surrogate evaluation spaces using their respective populations. In an $m$-objective optimization problem, these populations are defined independently for each space:
\begin{equation}
\begin{aligned}
\mathcal{P}_{\text{true}} = \Big\{ \left( \bm{x}_i, \bm{y}_i \right) \,\Big|\,\ 
&\bm{y}_i = \bm{f}(\bm{x}_i),\ 
i = 1, \dots, n_{\text{true}} \Big\}, \\
\mathcal{P}_{\text{sur}} = \Big\{ \left(\bm{x}_i, \hat{\bm{y}}_i, \hat{\boldsymbol{\sigma}}_i \right) \,\Big|\,
\ 
&\hat{\bm{y}}_i = \hat{\bm{f}}(\bm{x}_i),\  
i = 1, \dots, n_{\text{sur}} \Big\},
\end{aligned}
\end{equation}
where $\bm{x}_i \in \mathcal{X}\subset \mathbb{R}^d,$ and $\bm{y}_i = \bm{f}(\bm{x}_i) \in \mathbb{R}^{m}$. In the surrogate evaluation space, $\hat{\bm{y}}_i = \hat{\bm{f}}(\bm{x}_i) \in \mathbb{R}^{m}$ denotes the surrogate model’s prediction for the $i\text{-th}$ solution, and $\hat{\boldsymbol{\sigma}}_i \in \mathbb{R}^{m}_+$ indicates the corresponding predictive uncertainty. Notably, the true evaluation space contains all solutions evaluated by the true objectives, while the surrogate evaluation space consists of candidate solutions generated at the current step and evaluated by surrogate models.

\subsection{TabPFN}
This work is the first to adopt TabPFN (Tabular Prior-Data Fitted Network) as a surrogate model for optimization. TabPFN is a transformer-based probabilistic estimator that enables high-accuracy, low-latency one-shot inference without requiring task-specific training \cite{a28,a29,a30}, making it highly suitable for expensive black-box optimization. TabPFN outputs a discrete probability distribution over predefined value intervals. Specifically, for a candidate solution $\bm{x}_i \in \mathbb{R}^d$, the model returns a probability vector:
\begin{equation}
\bm{p}_i = \text{TabPFN}(\bm{x}_i)=[p^{(i)}_{1},\cdots p^{(i)}_{K}]\in [0,1]^{K},
\end{equation}
where $\sum_{k=1}^{K} p^{(i)}_{k} = 1$, $K$ denotes the number of predefined value intervals, and \( p_k^{(i)} \) represents the predicted probability that the objective value of the \( i \)-th solution falls into the \( k \)-th interval.

\section{Methodology}

\subsection{Overview}

\begin{algorithm}[tb]
\caption{DB-SAEA}
\label{alg:algorithm}
\textbf{Input}: Multi-objective black-box function $\bm{f}(\cdot)$, evaluation budget $FE_\text{max}$, ELA module $\Lambda_{\theta_\Lambda}$, mata-policy $\pi_{\theta_\pi}$\\
\textbf{Output}: Evaluated solutions $\mathcal{P}_\text{true}$
\begin{algorithmic}[1] 
\STATE $X\gets$ Generate an initial set of $N$ solutions using LHS
\STATE $Y\gets\bm{f}(X)$
\STATE Set $\mathcal{P}_\text{true}\gets(X,Y)$
\STATE $\hat{\bm{f}}(\cdot)\gets\bm{f}(\cdot)$ is approximated by TabPFNs
\STATE Set evaluation counter $t\gets N$
\WHILE{$t<FE_\text{max}$}
\STATE Generate offspring population $X_\text{off}$
\STATE $\mathcal{P}_\text{sur}\gets (X_\text{off},\hat{\bm{f}}(X_\text{off}),\hat{\bm{\sigma}}_\text{off})$ 
\STATE Landscape-aware state $\bm{z}_t \gets \Lambda_{\theta_\Lambda}(\mathcal{P}_\text{true},\mathcal{P}_\text{sur})$
\STATE Optimization state $\bm{s}_t \gets [\bm{z}_t, t/FE_\text{max}]$
\STATE Obtain an action $ a_t \gets \pi_{\theta_{\pi}}(\bm{s}_t)$
\IF {$a_t\neq a_1$}
\STATE Select elite solutions $X^*\subset X_\text{off}$ based on the infill criterion specified by $a_t$
\STATE ${Y}^*\gets{\bm{f}}(X^*)$
\STATE $\mathcal{P}_\text{true}\gets \mathcal{P}_\text{true} \cup (X^*,{Y}^*)$ and update $\hat{\bm{f}}(\cdot)$
\STATE $t\gets t+|X^*|$
\ENDIF
\ENDWHILE
\end{algorithmic}
\end{algorithm}
We propose a novel MetaBBO framework, DB-SAEA (see Algorithm 1 and Figure 2), where a SAEA serves as the low-level optimizer. Initially, solutions generated via Latin Hypercube Sampling (LHS) are evaluated using true objective functions $\bm{f}(\cdot)$, and these are used to construct multiple TabPFN surrogate models. Candidate solutions are then generated using a hybrid sampling strategy that integrates NSGA-III \cite{a31}, CDM-PSL \cite{a32}, and qNEHVI \cite{a33}, and are evaluated by the surrogate models. A bi-space ELA $\Lambda_{\theta_\Lambda}$ constructs landscape-aware representations from both true and surrogate evaluation spaces, These representations are provided to the meta-policy \( \pi_{\theta_\pi} \), which determines whether to proceed with true evaluations or discard the candidates and resample. If the candidates are selected, the meta-policy chooses an infill criterion and identifies elite solutions for true evaluation by \( \bm{f}(\cdot) \), followed by surrogate model updates. This process repeats until the evaluation budget is exhausted.

\subsection{Bi-space Neural Network-based ELA}

The ELA module of DB-SAEA first processes decision variables and objective values through a Population Information Embedding (PIE) module, which includes normalization and generates structured input representations. It then proceeds in two stages, where each stage separately applies a Ts-Attn block to both the true and surrogate evaluation spaces, enabling the extraction of informative features from each space independently.

\subsubsection{Population Information Embedding}
PIE module first applies min-max normalization to both $\mathcal{P}^t_\text{true}$ and $\mathcal{P}^t_\text{sur}$ at time step $t$. Specifically, for the decision vectors $\{\bm{x}_{i}\}_{i=1}^{n_\text{true}}$ and $\{\bm{x}_i\}_{i=1}^{n_\text{sur}}$, normalization is performed against the search range. For the objective values $\{\bm{y}_{i}\}_{i=1}^{n_\text{true}}$, $\{\hat{\bm{y}}_{i}\}_{i=1}^{n_\text{sur}}$, and the surrogate uncertainties $\{\hat{\bm{\sigma}}_{i}\}_{i=1}^{n_\text{sur}}$, normalization is based on the extremum values observed in $\mathcal{P}^t_\text{true}$ and $\mathcal{P}^t_\text{sur}$, respectively. This ensures unified representation and generalization by scaling all values to $[0,1]$. For a $d$-dimensional optimization problem with $m$ objectives, the normalized observations from $\mathcal{P}^t_\text{true}$ and $\mathcal{P}^t_\text{sur}$ are then reorganized: 
\begin{equation}
\begin{aligned}
M^{t}_\text{true}=\{\{\{{x}^t_{i,j},y^t_{i,obj}\}_{i=1}^{n_{\text{true}}}\}_{j=1}^d\}_{obj=1}^m,\\ 
M^{t}_\text{sur}=\{\{\{{x}^t_{i,j},\hat{y}^t_{i,obj},\hat{\sigma}^t_{i,obj}\}_{i=1}^{n_{\text{sur}}}\}_{j=1}^d\}_{obj=1}^m,
\end{aligned}
\end{equation}
where $M^t_\text{true}\in\mathbb{R}^{m\times d\times n_\text{true}\times 2}$ and $M^t_\text{sur}\in\mathbb{R}^{m\times d\times n_\text{sur}\times 3}$. Then, $M^t_\text{true},M^t_\text{sur}$ linearly projected via embedding matrices $W^\text{true}_\text{emb}\in \mathbb{R}^{2\times h}$ and $W^\text{sur}_\text{emb} \in \mathbb{R}^{3\times h}$ to obtain $E^t_\text{true},E^t_\text{sur}\in\mathbb{R}^{m\times d\times n \times h}$. Here, $h$ represents the hidden dimension.

\subsubsection{Stage One: Multi-objective Feature Extraction} 
In the first stage, the model extracts per-objective, individual-level landscape representations from both the true and surrogate spaces using two Ts-Attn blocks, each applied across all objectives within its respective space. The Attn block adopts the Transformer architecture \cite{a34} with layer normalization \cite{a35} replacing batch normalization \cite{a36}. For the $obj$-objective, Ts-Attn takes $E^t_{obj} \in \mathbb{R}^{d \times n \times h}$ and enables information interaction across both individuals and dimensions. \textbf{1) Cross-individual attention:} For each evaluation space, we apply an Attn block to allow solutions within the respective population to exchange information along the same dimension. \textbf{2) Cross-dimension attention:} In each evaluation space, the Cross-individual attention yields $m$ representations, one for each objective. We then apply a second Attn block to these representations to enable information exchange across dimensions within each solution. This involves transposing the output to shape $n\times d \times h$ and injecting positional encodings to preserve the dimensional order of each solution. \textbf{3) Pooling over problem dimensions:} For each objective, the output from the Ts-Attn module is aggregated via mean pooling along the dimension axis (i.e., over problem variables). The resulting are then stacked across all objectives to construct the multi-objective, individual-level landscape representations $\bm{S}^t_\text{true}\in\mathbb{R}^{m\times n_\text{true}\times h}$ and $\bm{S}^t_\text{sur}\in\mathbb{R}^{m\times n_\text{sur}\times h}$.

\subsubsection{Stage Two: Cross-objective Feature Aggregation} 
The representations $\bm{S}^t_\text{true}$ and $\bm{S}^t_\text{sur}$ are further refined using two additional Ts-Attn blocks, each comprising a Cross-individual and a Cross-objective attention mechanism. Unlike Stage One, we perform mean pooling along both the individual and objective axes to extract unified global representations, resulting in $\bm{s}'^{\;t}_\text{true}, \bm{s}'^{\;t}_\text{sur} \in \mathbb{R}^{h}$. These two vectors are then concatenated into a unified $2h\text{-dimensional}$ landscape-aware representation $\bm{z}_t = [\bm{s}'^{\;t}_\text{true},\bm{s}'^{\;t}_\text{sur}]\in\mathbb{R}^{2h}$.
The highly parallelizable attention-based architecture ensures that DB-SAEA remains scalable with respect to problem dimensionality, population size, and the number of objectives. Additional details are provided in \textbf{Appendix A.1}. 
\subsection{Low-level Optimizer Based on SAEA}
\subsubsection{Surrogate Model} 
For each objective function $f_{obj}$, we set an independent TabPFN model as a surrogate. Given a solution $\bm{x}_i$, the model returns a discrete probability distribution over $K$ bins defined by edges $\{b_k\}_{k=0}^{K}$. Let $\{p_k^{(i,obj)}\}_{k=1}^{K}$ denote the predicted probabilities corresponding to each bin for the $i$-th solution on the $obj$-th objective. To compute the scalar prediction and its uncertainty for the $obj\text{-th}$ objective, we define the midpoint of each bin as $\mu_k = (b_{k-1} + b_k)/2$ and estimate the predicted value and standard deviation of $\bm{x}_i$ using the following equations:
\begin{align}
\begin{aligned}
\hat{y}_{i,obj} = \sum_{k=1}^{K} p_k^{(i,obj)} \cdot \mu_k, \\
\hat{\sigma}_{i,obj} = \sqrt{ \sum_{k=1}^{K} p_k^{(i,obj)} (\mu_k - \hat{y}_{i,obj})^2 }.
\end{aligned}
\end{align}

\subsubsection{Evolutionary Algorithm} 
In the evolutionary algorithm, we incorporate multiple candidate generation strategies, including NSGA-III, CDM-PSL, and qNEHVI. The effectiveness of hybrid sampling has been validated in prior studies \cite{a32,a46,a47}. The hybrid design in DB-SAEA, combining NSGA-III for global exploration with CDM-PSL and qNEHVI for focused exploitation, enhances both the diversity and structural richness of candidate solutions.

It is worth noting that only NSGA-III is used for candidate generation during training, while the hybrid strategy is applied in testing. Since CDM-PSL and qNEHVI incur high computational costs, they are unsuitable for repeated training use. Moreover, DB-SAEA focuses on controlling candidate selection based on the optimization state rather than configuring the evolutionary algorithm. Experimental results confirm that training with NSGA-III alone is sufficient for strong test performance. Details of the hybrid strategy are provided in \textbf{Appendix A.2}.

\subsubsection{Infill Criterion} 
In designing the infill criteria, we draw inspiration from the EIC-MSSAEA method \cite{a43}. Specifically, we employ five different criteria: ND-A, two variants of $\text{ND-}\Delta PBI$ that respectively emphasize convergence and diversity, and two types of EPDI that focus on exploration and exploitation, respectively. Detailed definitions are provided in \textbf{Appendix A.3}.
\subsection{Model the Evolution Search Procedure as an MDP}
To enable dynamic control over the surrogate-assisted multi-objective optimization process, we model the search as a discrete-time, finite-horizon Markov Decision Process (MDP), defined as $\mathcal{M}=(\mathcal{S},\mathcal{A},{P},{r},\gamma )$, where $\gamma$ denotes discount factor. \textbf{State Space}. $\mathcal{S}$ denotes the state space that reflects optimization status. At each decision step $t$, the state vector $\bm{s}_t=[\bm{z}_t,\, \rho_t] \in \mathbb{R}^{2h+1}$ combines a landscape representation $\bm{z}_t \in \mathbb{R}^{2h}$ from a two-stage ELA module and $\rho_t = t / {FE}_{\max}$ is a normalized scalar indicating the proportion of the evaluation budget that has been consumed up to step $t$. \textbf{Action Space}. Define action space $\mathcal{A}=\{{a_1},a_2,\cdots,a_6\}$, where $a_1$ represents regenerating new candidate solutions, while others trigger true evaluations of elite individuals, with each action corresponding to one of the five predefined infill criteria. \textbf{Transition Dynamics}. The state transition ${P}$ can be expressed as a conditional probability $p(\bm{s}_{t+1}|\bm{s}_t,a_t)$, but this distribution is not explicitly available due to the black-box nature of the optimization process. Nevertheless, this does not hinder learning, as our model-free reinforcement learning approach relies only on sampled trajectories, without requiring knowledge of the transition probabilities. \textbf{Reward Function}. The reward function ${r}(s_t, a_t)$ provides feedback based on the impact of the selected action on optimization progress. If the agent continues surrogate evolution $(a_t=a_1)$, it receives a neutral reward of 0. When true evaluation is triggered $(a_t\in \{a_2,\cdots,a_6\})$, the reward reflects whether the newly evaluated solutions improve the Pareto front. Formally:
\begin{align}
r_t = 
\begin{cases}
0, & a_t = a_1 \\
1.0 + \lambda \cdot \sum_{i=1}^{k} \dfrac{d_i}{d^{(i)}_{\text{ref}}}, & \text{if the front is improved} \\
-1.0, & \text{otherwise.}
\end{cases}
\end{align}
Here, $k$ denotes the number of newly selected candidate solutions for true evaluation. For each solution, $d_i$ represents the Manhattan distance to the closest point on the previous Pareto front (before improvement), and ${d^{(i)}_{\text{ref}}}$ is the Manhattan distance from that closest point to the origin. This normalization ensures consistency across tasks with different scales and enables a more stable learning signal.

We formulate the meta-level objective of DB-SAEA as a bi-level optimization problem. The lower level uses a surrogate-assisted optimizer $\mathcal{O}$, while the upper level learns a meta-policy $\pi_{\theta_\pi}$ to guide the search across tasks $\mathcal{T}_i\sim \mathcal{D}$. At each step $t$, the state $\bm{s}_t$ is constructed by combining the bi-space landscape representation $\Lambda_{\theta_{\Lambda}}(\mathcal{O}, \mathcal{T}_i)$, extracted via the ELA module of DB-SAEA, and a progress scalar. Based on this state, the meta-policy selects an action $a_t = \pi_{\theta_\pi}(\bm{s}_t)$. The overall training objective is to maximize the expected performance across tasks:
\begin{align}
\begin{aligned}
J(\theta) = \mathbb{E}_{\mathcal{T} \sim \mathcal{D}} \left[ R(\mathcal{O}, \pi_{\theta_\pi}, \mathcal{T}) \right] \approx \frac{1}{N} \sum_{i=1}^{N} \sum_{t=1}^{T} \text{perf}(\mathcal{O}, a_t, \mathcal{T}_i),
\label{eq1}
\end{aligned}
\end{align}
where $\text{perf}(\cdot)$ is a task-specific performance metric, $N$ is the total number of training tasks, and $R(\cdot)$ represents total reward accumulated by following the meta-policy.

\subsection{Training Method}
As formalized in Equation \eqref{eq1}, our objective is to maximize the expected cumulative performance achieved by the collaboration of the surrogate-assisted optimizer $\mathcal{O}$ and the meta-policy $\pi_{\theta_\pi}$ across multiple tasks $\mathcal{T}_i \sim \mathcal{D}$. To enable transferability across diverse optimization problems, DB-SAEA explicitly considers the cross-task transferability of the meta-policy. We adopt a \textbf{parallel sampling} and \textbf{centralized training} paradigm, implemented via an offline reinforcement learning framework based on Dueling DQN \cite{a44}. Multiple multi-objective optimization environments (i.e., tasks) interact in parallel with the current meta-policy to generate heterogeneous state-action-reward trajectories. These experiences are aggregated into a centralized replay buffer and used to jointly train a meta-policy with shared parameters. This sampling-training loop is iterated until convergence. Notably, the bi-space ELA module is co-trained with the meta-policy during the reinforcement learning process, enabling tight integration between landscape-aware representation learning and adaptive decision-making.

To enable parallel sampling, we employ \textbf{Ray} \cite{a37}, an open-source framework for parallel processing in machine learning applications. With Ray, the sampling tasks can be distributed across multiple CPUs and GPUs, allowing simultaneous interaction with multiple environments.

\section{Experimental Studies}
In this section, we aim to address the following research questions: \textbf{RQ1}: How stable and convergent is the training process of the meta-policy across multiple optimization environments? \textbf{RQ2}: How well does the proposed approach transfer to new tasks with higher-dimensional problem settings? \textbf{RQ3}: What are the advantages of using TabPFN over GP in the context of optimization? \textbf{RQ4}: What is the impact of controlling only the evolutionary algorithm or only the infill criterion, compared to the proposed dual-control strategy? Additional experimental results are provided in \textbf{Appendix B}.
\subsection{Experimental Setup}
\subsubsection{Benchmark Problems and Baselines}
To comprehensively validate the performance of DB-SAEA, experiments were conducted on nine benchmark problems, 2- and 3-objective ZDT1-3 \cite{a38} and DTLZ2-7 \cite{a39}. Moreover, we compare DB-SAEA against multiple state-of-the-art and classical algorithms, including CDM-PSL \cite{a32}, MOEA/D-EGO \cite{a40}, USeMO-EI \cite{a41}, qNEHVI \cite{a33}, and NSGA-II \cite{a42}. Additionally, we include an ablation variant, denoted as DB-SAEA-NSGA, in which only NSGA-III is used to generate candidate solutions, thereby removing the hybrid candidate generation strategy from the original DB-SAEA framework.
\subsubsection{Training and Testing Settings}
We employ a parallel sampling strategy to interact with multiple benchmark environments simultaneously. All experiences are stored in a centralized replay buffer for joint training of the meta-policy and bi-space ELA module. The buffer is cleared and refreshed after each training round. The training batch size is set to 64. We use the Adam optimizer with an initial learning rate of 0.0001. To assess transferability, we apply a \textbf{leave-one-task-out cross-validation} strategy. In each run, we train the meta-policy and ELA module on eight tasks and then test them on the remaining unseen task, which has a higher problem dimensionality. This process is repeated 9 times, ensuring that each task is used once as the test task. Moreover, each test round is independently repeated 10 times. During both training and testing phases, we initialize the population in the true evaluation space with 80 solutions, while setting the remaining evaluation budget to 40 for use during the optimization process. At each decision step, one solution is selected for true evaluation. All attention modules within the bi-space ELA architecture employ a single-head configuration with hidden dimension $h=16$.

All experiments were conducted on a compute cluster equipped with 4 × 48GB VGPUs and 4 × 25-core Intel Xeon Platinum 8481C processors. With the Ray framework, we achieved efficient task-level parallel sampling and centralized training, reducing each sampling-training cycle to around 5 minutes.

\subsubsection{Evaluation Metric}
We employ the hypervolume (HV) metric \cite{a48} to evaluate the quality of the obtained solution sets. A higher HV value reflects better better performance.

\subsection{Training Stability and Convergence (RQ1)}

\begin{figure}[t]
\centering
\includegraphics[width=0.9\columnwidth]{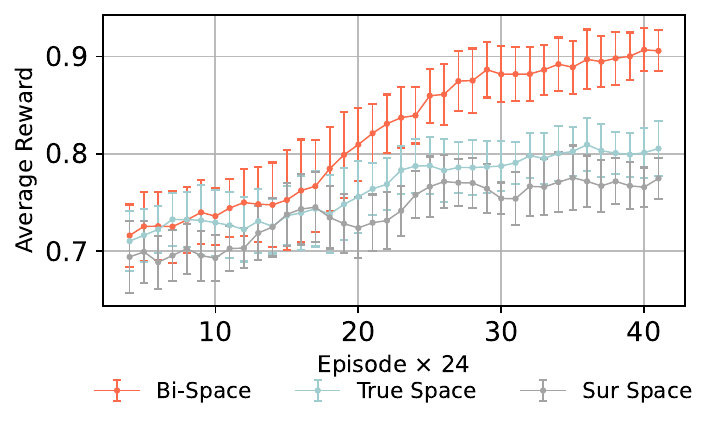} 
\caption{Average reward per true evaluation}
\label{fig3}
\end{figure}

\begin{figure*}[tb]
  \centering
  \begin{minipage}[t]{0.465\linewidth}
    \centering
    \includegraphics[width=\linewidth]{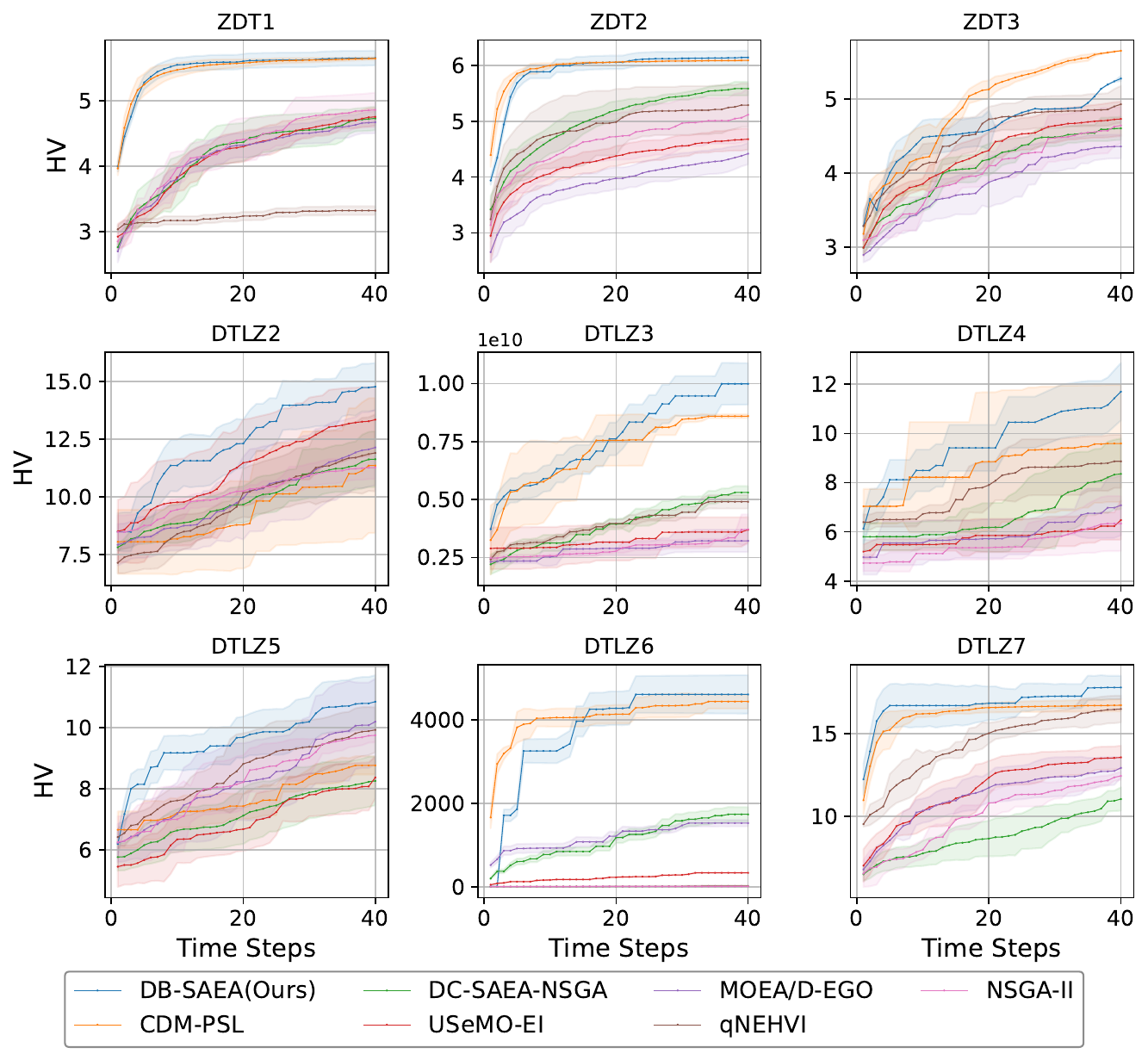}
    \caption{Performance comparison on unseen 30D tasks}
    \label{fig:fig1}
  \end{minipage}
  \hfill
  \begin{minipage}[t]{0.53\linewidth}
    \centering
    \includegraphics[width=\linewidth]{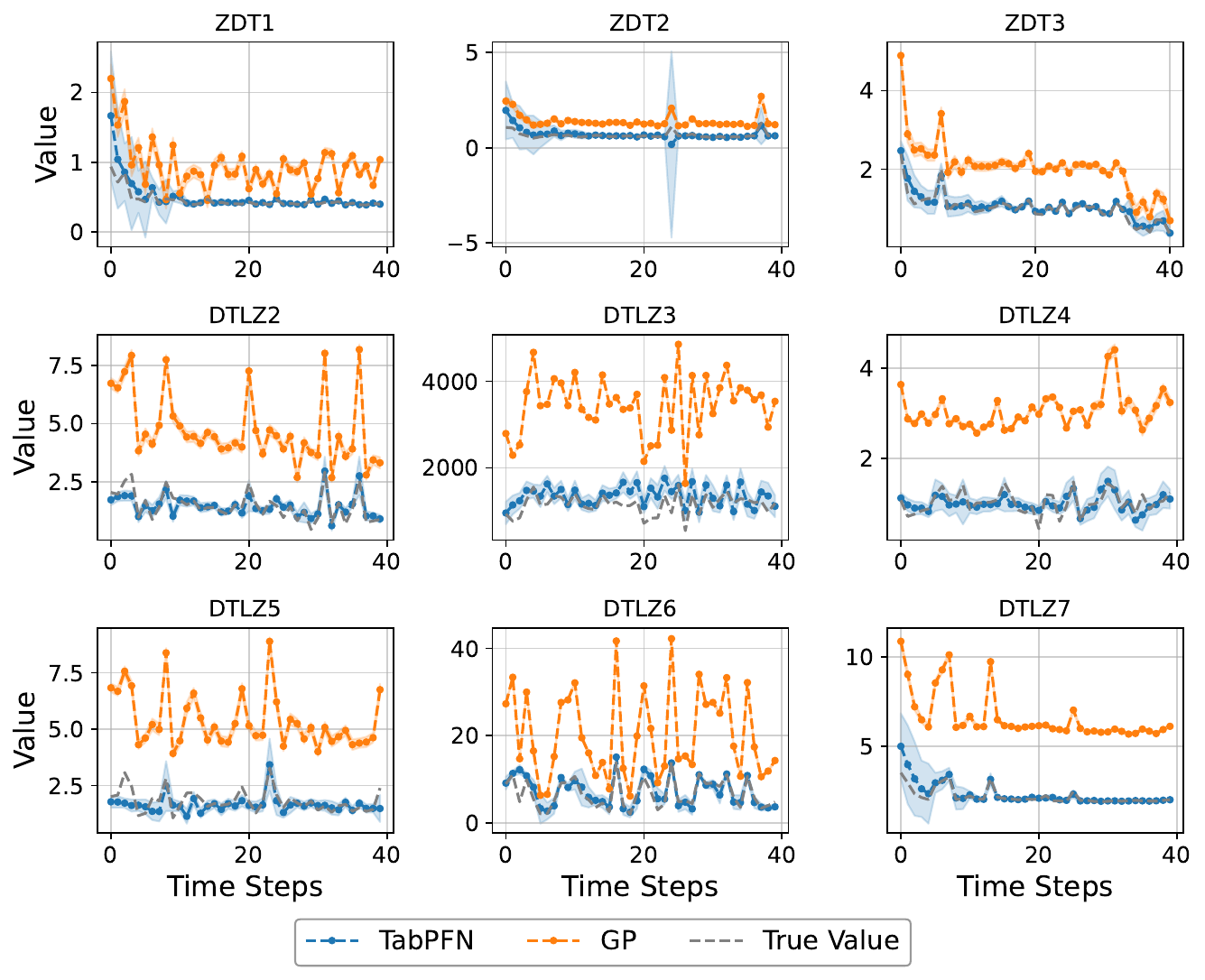}
    \caption{Prediction and uncertainty of surrogate models}
    
    \label{fig:fig2}
  \end{minipage}
\end{figure*}

\begin{figure*}[tb]
\centering
\includegraphics[width=0.80\linewidth]{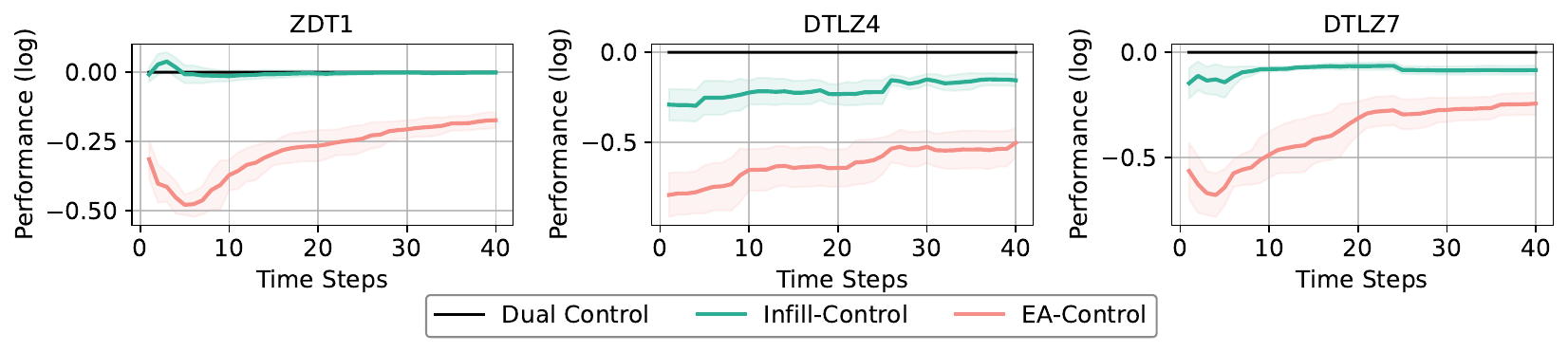} 
\caption{Performance comparison of single and dual control}
\label{fig6}
\end{figure*}

We conduct 9 training runs using a leave-one-task-out cross-validation strategy. In each run, eight optimization environments are used for training, and each environment is configured with three different dimensional settings (15, 20, and 25), resulting in a total of 24 training tasks. Figure 3 presents the average reward per true evaluation, obtained by averaging across environments at each training episode. The curve is smoothed using a moving average with a window size of 5. We compare three ELA variants during training: Bi-Space (features from both true and surrogate evaluations), True Space (only true-evaluation features), and Sur Space (only surrogate-based features). Bi-Space achieves higher rewards, demonstrating that combining accurate but sparse true data with low-cost surrogate information enhances optimization guidance. Vertical bars indicate ±1 standard deviation over 9 runs.

\subsection{Zero-shot Performance (RQ2)}

To assess zero-shot transferability, we evaluate DB-SAEA on unseen tasks with higher dimensionality. After training on eight benchmark tasks, the model is directly tested on a held-out task with 30 decision variables. Figure 4 shows that DB-SAEA generally outperforms all baseline methods. In particular, DB-SAEA significantly outperforms its ablated variant DB-SAEA-NSGA, validating the benefit of the hybrid candidate generation strategy. Notably, although DB-SAEA adopts candidate generation components from CDM-PSL and qNEHVI, the incorporation of dual control yields a substantial performance gain beyond these individual components. While DB-SAEA performs comparably to CDM-PSL on simpler problems like ZDT1 and ZDT2, it shows marked improvements on more complex tasks.

\subsection{Comparing TabPFN with Gaussian Process (RQ3)}

Figure 5 compares the predictive performance of TabPFN and GP on candidate solutions selected for true evaluation across nine 30-dimensional benchmark tasks. At each true evaluation step, the figure shows the mean predicted values across all objectives, along with the corresponding mean true objective values for the selected solutions. The blue line represents the multi-objective mean prediction produced by TabPFN, the orange line indicates that of the GP model, and the gray line shows the actual mean of the true objective values. Shaded areas denote the ±1 standard deviation around each model’s predictions, reflecting prediction uncertainty. It can be observed that TabPFN yields predictions that are overall closer to the true values, with most true values falling within its ±1 std range, whereas the GP model displays more noticeable deviations.

\subsection{Comparison of Single and Dual Control (RQ4)}

In this experiment, we construct two ablation variants to evaluate the effectiveness of the proposed dual-control mechanism of DB-SAEA: Infill Control, which only governs the selection of infill criteria, and EA Control, which only decides whether to perform true evaluation or continue surrogate-assisted search. Figure 6 presents a performance comparison across control strategies using the logarithmic hypervolume ratio computed with base 2. DB-SAEA with dual control achieves notably better performance on complex tasks such as DTLZ4 and DTLZ7, while showing comparable results to Infill Control on simpler problems like ZDT1. Evidently, Infill Control outperforms EA Control, suggesting that dynamic control over infill criterion selection contributes more significantly to optimization performance. 
\section{Conclusion}
In this paper, we propose DB-SAEA, a MetaBBO framework tailored for expensive multi-objective optimization problems, which learns a meta-policy to jointly control evolutionary algorithm and infill criterion selection. It integrates a bi-space ELA module to capture optimization states from both true and surrogate evaluation spaces, while ensuring scalability across problem dimensions, population sizes, and objectives. In addition, we employ TabPFN as a surrogate model to provide accurate predictions and uncertainty estimates without requiring task-specific training. Experiments show that DB-SAEA achieves strong performance, particularly on complex tasks, and demonstrates effective zero-shot transfer capability. Ablation studies validate the effectiveness of both the dual-control mechanism and the bi-space landscape modeling. Moreover, TabPFN significantly outperforms Gaussian Processes across all benchmark tasks. Future work may explore three directions: 1) expanding beyond dual control to incorporate more optimization decisions, such as surrogate model selection or evolutionary algorithm configuration, to enhance adaptive capability; 2) enhancing the joint modeling of true and surrogate landscape features by integrating them through advanced mechanisms such as cross-attention; and 3) further improving DB-SAEA for more complex scenarios, such as constrained multi-objective or dynamic optimization problems, to advance meta-level control in BBO.

\section{Acknowledgments}
This research was supported by the National Natural Science Foundation of China (Grant No. 72301286, 72431011).

\bibliography{aaai2026}

\end{document}